\documentclass{article}

\usepackage{arxiv}

\usepackage[utf8]{inputenc} 
\usepackage[T1]{fontenc}    
\usepackage[]{hyperref}
\usepackage{url}            
\usepackage{booktabs}       
\usepackage{amsfonts}       
\usepackage{nicefrac}       
\usepackage{microtype}      
\usepackage{cleveref}       
\usepackage{lipsum}         
\usepackage{graphicx}
\usepackage{subcaption}
\usepackage{natbib}
\usepackage{doi}
\usepackage{multirow}

\title{Evaluation of ONCOTIMIA: an LLM‑based system for supporting tumour Boards }

\newif\ifuniqueAffiliation
\uniqueAffiliationtrue

\author{%
	Luis Lorenzo$^{1}$,
	Marcos Montaña-Méndez$^{1}$,
	Sergio Figueiras$^{1}$,
	Miguel Boubeta$^{1,*}$,
	Cristóbal Bernardo-Castiñeira$^{1,*}$ \\
	$^{1}$ Innovation Department, Bahía Software SLU, Ames (A Coruña), Spain \\
	$^{*}$ Corresponding author(s): 
	\texttt{miguel.boubeta@bahiasoftware.es, cristobal.bernardo@bahiasoftware.es}
}
\renewcommand{\headeright}{}
\renewcommand{\undertitle}{Technical Report}
\renewcommand{\shorttitle}{}

\hypersetup{
	pdftitle={A template for the arxiv style},
	pdfsubject={q-bio.NC, q-bio.QM},
	pdfauthor={David S.~Hippocampus, Elias D.~Striatum},
	pdfkeywords={First keyword, Second keyword, More},
}

\graphicspath{{./figures/}}

\begin{document}
\maketitle

\begin{abstract}
Multidisciplinary tumour boards (MDTBs) play a central role in oncology decision-making but require manual processes and structuring large volumes of heterogeneous clinical information, resulting in a substantial documentation burden. In this work, we present ONCOTIMIA, a modular and secure clinical tool designed to integrate generative artificial intelligence (GenAI) into oncology workflows and evaluate its application to the automatic completion of lung cancer tumour board forms using large language models (LLMs). The system combines a multi-layer data lake, hybrid relational and vector storage, retrieval-augmented generation (RAG) and a rule-driven adaptive form model to transform unstructured clinical documentation into structured and standardised tumour board records. We assess the performance of six LLMs deployed through AWS Bedrock on ten lung cancer cases, measuring both completion form accuracy and end-to-end latency. The results demonstrate high performance across models, with the best performing configuration achieving an $80\%$ of correct field completion and clinically acceptable response time for most LLMs. Larger and more recent models exhibit best accuracies without incurring prohibitive latency. These findings provide empirical evidence that LLM- assisted autocompletion form is technically feasible and operationally viable in multidisciplinary lung cancer workflows and support its potential to significantly reduce documentation burden while preserving data quality. 
\end{abstract}

\keywords{GenAI \and LLMs \and Vector database \and Embeddings \and RAG \and Tumour boards \and Lung cancer form autocompletion}

\section{Introduction}\label{sec:introduction}
In recent years, advances in transformer-based architectures have firmly established LLMs as a foundational technology in biomedical informatics. Early developments in general-purpose LLMs (e.g., GPT-3 and its successors) revealed emergent capabilities in clinical summarisation, question answering, report generation, coding support and contextual reasoning \citep{brown2020}. Subsequent work has shown that domain-adapted models, such as BioGPT \citep{luo2022}, BioMedLM, PubMedBERT \citep{gu2021} and Med-PaLM \citep{singhal2023}, achieve expert-level performance on diverse medical reasoning benchmarks. An expanding evidence base further demonstrates that LLMs can reliably extract salient clinical information and support guideline-informed recommendations when deployed with appropriate safeguards. Recent prospective evaluations in hospital settings indicate that LLM-assisted clinical documentation can meaningfully reduce clinician workload while maintaining high linguistic quality \citep{bracken2025, nori2023}. Nevertheless, these systems continue to require rigorous oversight owing to risks of hallucinations, incomplete contextualisation and occasional misinterpretation of clinical guidelines. 

In medicine, multidisciplinary management (MDM) offers cancer patients the advantage of having specialists from different medical fields collaboratively involved in treatment planning. This approach is usually implemented through multidisciplinary clinics, such as breast units, where various experts assess patients, perform physical examinations, request and conduct diagnostic tests efficiently, and jointly evaluate potential treatment options. MDM is also conducted through multidisciplinary tumour board (MDTB) meetings, which are structured sessions in which all relevant patient information is collected, and key specialists convene to discuss the diagnosis and management of cancer patients \citep{elsaghir2014}. However, studies show that clinicians spend significant effort managing and analysing information within electronic health records (EHRs), reducing the time for direct patient care, especially in high-complexity fields such as oncology \citep{arndt2017}. MDTBs are particularly affected because each case requires the preparation of standardised summaries, structured staging information, pathology details, radiological interpretations, biomarker data, allergy profiles and records of prior treatment, demands that are specially challenging in settings with limited staff resources. The fragmentation of data across narrative notes, laboratory systems, pathology platforms and PACs often results in manual information retrieval and redundant re-entry of variables into MDTB case forms.

Automating pieces of this workflow is therefore both operationally compelling and clinically relevant. Early efforts using traditional natural language processing (NLP) methods demonstrated benefit in extracting structured information from radiology or pathology reports \citep{wang2018}. The transition from rule-based NLP to LLM-enabled generative systems, in addition to extracting information, also allows it to be synthesised into coherent drafts aligned with medical guidelines. 

Autocompletion in clinical documentation has emerged as a promising application of LLMs. Initial experiments in general EHR contexts have shown that LLMs can assist with automatic drafting of the main complaints, suggesting phrasing for assessment and plan sections, and generating templated texts conditioned on structured inputs \cite{ayers2023}. These systems have demonstrated that LLM-driven autocompletion improves efficiency and reduces repetitive typing. However, studies explicitly examining autocompletion for oncology MDTB forms remain extremely limited. To date, most published work in oncology has focused on information extraction (e.g., stage from clinical notes) or summarisation of radiology reports. Some recent pilot studies have explored the use of LLMs to generate oncology case summaries or harmonise staging descriptions \cite{chen2025}, but standardised autocompletion of MDTB forms, particularly in lung cancer, has not yet been rigorously evaluated in prospective settings. This represents a critical evidence gap given the structured, repetitive and data-dense nature of these forms and their importance in treatment decision making. 

Lung cancer has been one of the earliest and most active oncology domains for AI research due to the abundance of imaging, molecular data and clinical texts. NLP and deep learning models have been applied to staging extraction, biomarker result interpretation, automatic radiology summarisation and automated assessment of eligibility for targeted therapy or clinical trials \citep{esteva2019, hu2021, aldea2025}. 

In this work, our objective is to describe and technically evaluate the performance of ONCOTIMIA, a modular and secure LLM-based tool that integrates RAG and a rule‑driven adaptive form model to automate the completion of lung cancer tumour board forms. We assess its performance in a realistic tumour board setting, using synthetic but clinically representative cases. Six state‑of‑the‑art LLMs are evaluated in terms of form‑completion accuracy and end‑to‑end latency. Through this study, we aim to provide empirical evidence on the technical and operational feasibility of GenAI‑assisted autocompletion within oncology workflows, and to demonstrate its potential to reduce documentation burden while maintaining data quality. 

The following Section \ref{sec:oncotimia} introduces the ONCOTIMIA platform, outlining its architecture, data ingestion pipeline, and the design of lung cancer data schema. Section \ref{sec:materials} summarises the methodology for generating medical data records, the RAG workflow and the selection criteria for LLMs. Section \ref{sec:results} reports the performance evaluation results, and Section \ref{sec:conclusions} concludes by highlighting key findings, limitations, and directions for future work. 

\section{ONCOTIMIA tool description }\label{sec:oncotimia}
ONCOTIMIA is a modular system that integrates generative AI to support tumour board workflows and reduce documentation burden. Its core functionality focuses on the automatic autocompletion of standardised tumour board forms and the generation of structured patient summaries from heterogeneous clinical sources. The system also incorporates information retrieval and RAG‑assisted reasoning modules to facilitate case preparation; these features are intended to support review and do not replace clinical judgment. In this section, we present the system architecture and clinical data ingestion process and the definition of lung cancer form used in tumour boards committee workflows. 

\subsection{System architecture}\label{subsec:architecture}

\begin{figure}[!ht]
	\centering
	\includegraphics[width=14cm]{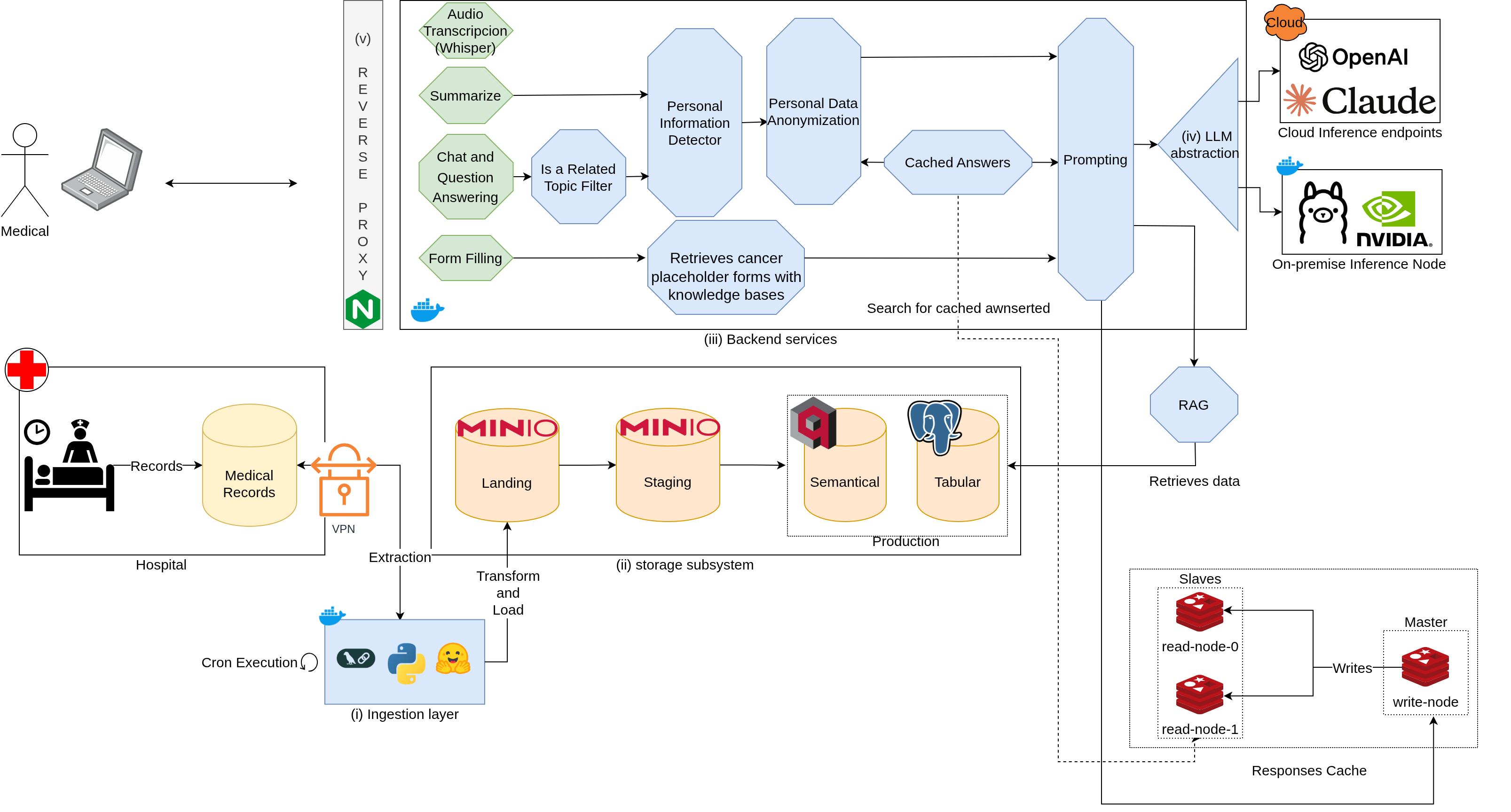}
	\caption{ONCOTIMIA architecture. (i) Data ingestion layer, (ii) storage subsystem, (iii) backend services, (iv) LLM abstraction layer and (v) reverse proxy.}
	\label{fig:fig1}
\end{figure}

The architecture of ONCOTIMIA has been conceived as a modular, scalable, and secure infrastructure designed to enable the seamless integration of GenAI into oncology workflows. Its design adheres to the principles of interoperability, traceability, maintainability, and controlled evolution, thereby ensuring long-term sustainability in complex clinical environments undergoing continuous technological transformation. From a conceptual standpoint, the ONCOTIMIA architecture (see Figure \ref{fig:fig1} for more details) is structured around a set of interconnected yet decoupled modules that operate in a coordinated manner through standardised interfaces and secure communication protocols. This modular arrangement enables the independent evolution of system components and the incremental incorporation of new functionalities without compromising overall system stability. 

Its design principles emphasise interoperability, traceability, maintainability and controlled evolution, ensuring robustness in complex clinical environments undergoing continuous technological change. The ONCOTIMIA architecture is structured into five core modules: (i) data ingestion layer, (ii) storage subsystem, (iii) backend services, (iv) LLM abstraction layer and (v) reverse proxy. 

The ingestion layer serves as the system’s entry point, acquiring and validating data from heterogeneous clinical sources, including electronic health records, structured and unstructured reports, laboratory results and administrative datasets. It implements automated pipelines for extraction, cleaning and normalisation to guarantee data quality and consistency prior to downstream processing. The use of mature data-engineering ecosystems enables reproducible, auditable and standard-compliant ingestion workflows, with native support for healthcare interoperability standards. The storage subsystem provides persistent, secure and traceable management of clinical data. ONCOTIMIA adopts a multilayered data lake structure comprising: (1) a landing layer for preserving source data in native formats, (2) a staging layer where data are transformed into analytically consistent representations, and (3) a refined layer optimised for high-performance queries and GenAI-driven retrieval tasks. The refined layer integrates both relational storages, suited for structured clinical variables, and vector databases supporting semantic search and embedding-based retrieval. The backend, built around a microservices paradigm, encapsulates the business logic required to support core tumour-board use cases. Key functionalities include: 

\begin{itemize}
	\item Automatic summarisation of clinical histories, combining structured and unstructured inputs.
	\item Autocompletion of tumour-specific forms, converting narrative text into semantically normalised representations aligned with oncology terminologies.
	\item A clinical assistant module, enabling question answering, hypothesis exploration and guideline-informed decision support. 
\end{itemize}

An intermediate abstraction layer mediates interactions between backend services and LLMs. It translates clinical requests into model-compliant queries, enforces safety and audit constraints and standardises output formatting. This layer enables model interchangeability and facilitates the integration of retrieval components and domain-specific knowledge bases. A reverse proxy manages traffic routing across system components. It enforces security policies, load balancing and rate limiting, while supporting real-time monitoring and audit logging. This layer ensures controlled and secure exposure of services to external and internal clients. 

\subsection{Data ingestion and ETL processes}\label{subsec:ingestion}
The clinical data ingestion process constitutes the entry point of the data processing pipeline and relies on a data lake architecture designed to efficiently and securely manage the heterogeneity of hospital information sources. This infrastructure supports the integration of both structured data (e.g., administrative records, laboratory results, demographic variables) and unstructured data (e.g., radiological reports, medical notes, clinical guidelines, oncology protocols, and supplementary documentation). The primary objective of this ingestion layer is to ensure complete preservation of the original content while enforcing quality-control mechanisms, format validation, and metadata generation to maintain full traceability of the information flow. 

The data lake is organised into three functional layers (landing, staging and refined) which reflect the progressive transformation of data from raw inputs to curated, analysis-ready outputs (see Figure \ref{fig:fig1} for more details). Documents received from hospital information systems or authorised external sources are stored in the landing layer, maintaining their original formats (e.g., .pdf, .docx or .txt) to preserve auditability and end-to-end traceability. 

Data stored in the landing layer feeds a set of sequential ETL (Extract, Transform and Load) processes implemented in Python, which constitute the operational backbone of the pipeline. First ETL process validates formats, applies integrity checks, and generates technical metadata during initial ingestion. The second ETL process extracts content and metadata from documents into the staging layer using specialised LangChain loaders (e.g., Docx2txtLoader and PyPDFLoader), followed by text cleaning (e.g., removal of line breaks or non-informative characters), tokenisation, lemmatisation, and stemming. The third and fourth ETL processes populate the refined layer by loading curated unstructured data into a vector storage system and structured patient information into a relational database. Unstructured text is encoded as semantic embeddings using the Nomic model and stored in a Qdrant vector store, enabling contextual RAG-based pipelines, and AI-assisted reasoning. On the other hand, structured and normalised clinical variables (e.g., demographics, coded diagnoses, tumour staging, and treatments) are stored in a PostgreSQL database, which serves as the analytical source of truth. The coexistence of relational and vectorial storage provides a hybrid integration of explicit clinical knowledge and contextual information derived from language models. 

\subsection{Lung cancer form schema}\label{subsec:form}
The MDTB lung cancer data form is organised into seven blocks, defined as logical units that group multiple questions. Transitions between blocks follow a non-linear, rule-based logic, whereby responses to specific questions determine the activation, omission, or redirection of subsequent sections. This adaptive structure allows the form to dynamically adjust to individual patient characteristics and the clinical context under evaluation. The form captures key clinical domains, including demographic information, smoking status and other risk factors, radiological and pathological findings, molecular biomarkers relevant to precision oncology, and prior treatments with their therapeutic intent. The overall structure is anchored by Block 1, which functions as the central node of the form. Block 1 consolidates core clinical variables (risk factors, comorbidities, ongoing medication, diagnostic test results, and detailed tumour profiling), and conditionally activates the remaining blocks based on pivotal responses. The main components of this block include: 
\begin{itemize}
	\item \textbf{Patient characteristics and medical history}: smoking status, comorbidities, allergies and prior malignancies.
	\item \textbf{Functional status and previous therapies}: ECOG performance score, radiotherapy or chemotherapy, and documented treatment refusals.
	\item \textbf{Imaging and endoscopic assessment}: free-text summaries of CT, PET-CT, bronchoscopy and other relevant procedures.
	\item \textbf{Histopathological and molecular analysis}: tumour histology, molecular biomarkers, PD-L1 expression, tumour mutational burden (TMB) and microsatellite status.
	\item \textbf{Tumour staging}: standardised categories capturing local, locoregional and systemic disease spread. 
\end{itemize}

Responses collected in Block 1 conditionally determine the activation of subsequent blocks collected in Table \ref{tab:table1}, enabling an adaptive and patient-specific data collection workflow. A reported history of malignancy activates Block 2 to capture details of earlier cancer diagnoses. Documentation of treatment refusal triggers Block 3 while indication of disease recurrence activates Block 4 to record affected sites. When a rebiopsy has been performed, Block 5 is enabled to document updated histology and molecular markers. A history of radiotherapy activates Block 6, which captures treatment intent, target lesions and timelines, whereas prior systemic therapy triggers Block 7 to document administered agents, therapeutic intent and treatment dates. This modular, conditionally driven design ensures that only clinically relevant information is collected for each case, resulting in a structured yet flexible representation of patient data. Such a context-aware data model supports downstream tasks including LLM-assisted summarisation, retrieval-augmented reasoning, and clinical decision support within multidisciplinary tumour board workflows. 

\renewcommand{\arraystretch}{1.3}
\begin{table}[!ht]
	\caption{Description of the content of Blocks 2 to 7 derived from Block 1.}
	\centering
	\begin{tabular}{rl}
		\toprule
		Block & Description \\
		\midrule
		2 & Previous neoplasms \\
		3 & Treatment refusal \\
		4 & Recurrence \\
		5 & Rebiopsy and new biomarkers \\
		6 & Radiotherapy \\
		7 & Chemotherapy \\
		\bottomrule
	\end{tabular}
	\label{tab:table1}
\end{table}

Table \ref{tab:table2} summarises the core clinical variables included in the lung cancer form, organised into thematic sections that reflect the logical structure of the oncological assessment process. Each section groups related fields according to their clinical meaning and functional role within the data model, while explicitly specifying the corresponding data type to ensure consistency, interpretability and suitability for downstream computational processing. 

The medical history section includes key baseline variables that characterise the patient’s background and prior clinical context. This section records smoking status encoded as a categorical variable with three possible states (smoker, non-smoker and ex-smoker), alongside binary indicators of previous neoplasia and documented treatment refusal. In addition, patient medication is represented as a categorical field, allowing for the structured documentation of the current medication (e.g., oral anticoagulation, antiplatelet agents, etc.). 

The performance status section captures the patient’s functional condition through the ECOG performance status core, encoded as an integer ranging from 0 (fully active) to 5 (dead). This variable is widely used in oncology to assess a patient’s ability to tolerate systemic treatments. 

\renewcommand{\arraystretch}{1.3}
\begin{table}[!ht]
	\caption{Subset of core clinical fields for the lung cancer form.}
	\centering
	\begin{tabular}{lll}
		\toprule
		Section & Field & Data type \\
		\midrule
		\multirow{4}{*}{Medical history} & Smoking status & Categorical \\
		& Previous neoplasia & Boolean \\
		& Treatment refusal & Boolean \\
		& Patient medication & Categorical \\
		\midrule
		Performance status & ECOG value & Integer \\
		\midrule
		\multirow{7}{*}{Diagnosis} & Local location & Categorical \\
		& Locoregional location & Categorical \\
		& Systemic location & Categorical \\
		& Hystology type & Categorical \\
		& Molecular marker & Categorical \\
		& PD-L1 value & Float \\
		& Recurrence & Boolean \\
		& Rebiopsy & Boolean \\
		\midrule
		\multirow{2}{*}{Treatment} & Radiotherapy & Boolean \\
		& Chemotherapy & Boolean \\	
		\bottomrule
	\end{tabular}
	\label{tab:table2}
\end{table}

The diagnosis section constitutes the most extensive component of the table and encompasses variables describing disease localisation, pathological characterisation and molecular profiling. Tumour extent is represented through categorical fields capturing local, locoregional and systemic involvement (e.g., lung, bone, liver, etc.). Histological tumour type is recorded as a categorical variable, enabling standardised classification of lung cancer subtypes (adenocarcinoma, squamous cell carcinoma, large cell carcinoma and small cell carcinoma). Molecular characterisation is incorporated through a categorical molecular marker field complemented by a continuous PD-L1 expression value that records quantitative immunohistochemical values, generally reported as the percentage of tumour cells expressing the marker. Molecular biomarkers are encoded as categorical variables that explicitly represent both the presence and absence of clinically actionable genomic alterations (e.g., EGFR, ALK, KRAS, BRAF, ROS1, etc.). The section also includes Boolean indicators for tumour recurrence and rebiopsy, which are essential for documenting disease evolution and the availability of updated pathological or molecular information. 

Finally, the treatment section records prior oncological interventions, specifically radiotherapy and chemotherapy, both encoded as Boolean variables. These fields provide a concise representation of previous treatment exposure and serve as key triggers for conditional workflows and more detailed treatment-specific documentation elsewhere in the system. 

\section{Materials and methods}\label{sec:materials}
Due to the highly sensitive nature of patient health data and the strict regulatory constraints governing its use, no real patient records were directly employed in the experimental evaluation of the proposed system. Instead, a fully synthetic dataset was constructed, using as a starting point a real clinical history that had been previously and irreversible anonymised in accordance with applicable data protection regulations. This reference case was used exclusively as a structural and narrative template, without retaining any real patient-identifiable or clinically traceable information. 

A total of ten Spanish synthetic clinical histories were generated, reflecting the real operational language of the clinical environment. Given the exploratory nature of this study and the need to assess the performance of different LLMs under controlled conditions, this initial experiment was intentionally limited in scope as a proof of concept to future large-scale deployment in real clinical settings. The synthetic cohort was generated using the Qwen3-14b LLM executed locally via the Ollama framework. This choice was motivated by the need to ensure full data governance and prevent any external data leakage. The model was prompted to produce multiple clinically plausible, internally consistent, and representative Spanish lung cancer patient histories that reflected the diversity of cases typically discussed in MDTBs, including variations in staging, molecular profiles, prior treatments, and clinical evolution. Following this initial generation phase, a two-step validation and refinement process was applied to ensure medical coherence and internal consistency. First, an automated reflection-based validation step was performed using GPT-OSS-120b model, which was tasked with critically reviewing each synthetic clinical history to detect logical inconsistencies, missing information, temporal contradictions, or medically implausible statements. The model was instructed to either confirm the coherence of the case or propose corrective revisions, which were then applied to the dataset. Second, the resulting synthetic cases were subjected to a final manual review by an expert in oncology, who assessed their clinical plausibility, internal consistency and suitability for use in a simulated tumour board setting. Only cases that passed this expert review were included in the final evaluation dataset. 

This multi-stage process ensures that the resulting dataset, preserves a high degree of clinical realism and complexity, making it suitable for a meaningful and rigorous assessment of the proposed system, based on a RAG architecture specifically designed for the automatic completion of structured lung cancer tumour board forms from unstructured clinical narratives in Spanish. The architecture integrates three main components: (i) document ingestion and preprocessing pipeline, (ii) a hybrid storage and retrieval layer, and (iii) an LLM-based generation layer. 

Clinical narratives are first segmented, normalised and embedded into a dense vector space using the Nomic embedding model. These embeddings are stored in a Qdrant database. At inference time, for each target form block, a query is constructed and use to retrieve the most relevant textual fragments from the vector storage. These retrieval contexts are then injected into a structured prompt template together with explicit instructions and the schema of the target form fields. This RAG-based approach also enables traceability, as each generated field can be linked back to the specific source fragments that supported it. 

Six LLMs models were selected for the experimental evaluation: GPT-OSS-20b, GPT-OSS-120b, Mistral-large-2402-v1, Pixtral-large-2502-v1, Qwen3-32b and Qwen3-next-80b. This set was designed to provide a representative and methodologically sound benchmark across different architectural families, parameter scales, and deployment profiles. GPT-OSS-20b and GPT-OSS-120b enable a controlled analysis of the impact of model scale within a single architectural lineage, isolating the effect of parameter count on extraction accuracy and reasoning stability. Mistral-Large-2402-v1 was included as a strong general-purpose model optimised for long-context understanding and complex reasoning, which is essential given the length and heterogeneity of the clinical narratives. Pixtral-Large-2502-v1 was selected for its strengths in structured generation and schema-constrained reasoning, which closely match the requirements of mapping free text to a predefined clinical form. Finally, Qwen3-32b and Qwen3-Next-80b were selected as high-performing open-weight models that combine strong predictive performance with operational feasibility, enabling reliable deployment in on-premise or tightly controlled infrastructures, as required in regulated clinical environments. 

Each synthetic clinical case was processed independently by the system, and each of the six models was used as the generation component within the same RAG pipeline, ensuring that all other components of the system remained strictly identical. This design isolates the effect of the language model itself on the quality and latency of the generated outputs. The automatically completed forms were then compared against the ground-truth structured information associated with each synthetic case. Field-level accuracy metrics were computed, and latency measurements were recorded end-to-end for each inference.

\section{Results}\label{sec:results}
The performance of the proposed form autocompletion tool was evaluated on the 10 synthetically generated lung cancer histories using the 6 selected LLMs available through AWS Bedrock (GPT-OSS-20b, GPT-OSS-120b, Mistral-large-2402-v1, Pixtral-large-2502-v1, Qwen3-32b and Qwen3-next-80b). For each LLM, we have assessed two dimensions: (i) accuracy, quantified as the percentage of correctly completed fields and (ii) model latency, measured as end-to-end response time (in seconds) per form. 

\begin{figure}[htbp]
	\centering
	\begin{subfigure}[b]{0.85\textwidth}
		\centering
		\includegraphics[width=0.85\linewidth]{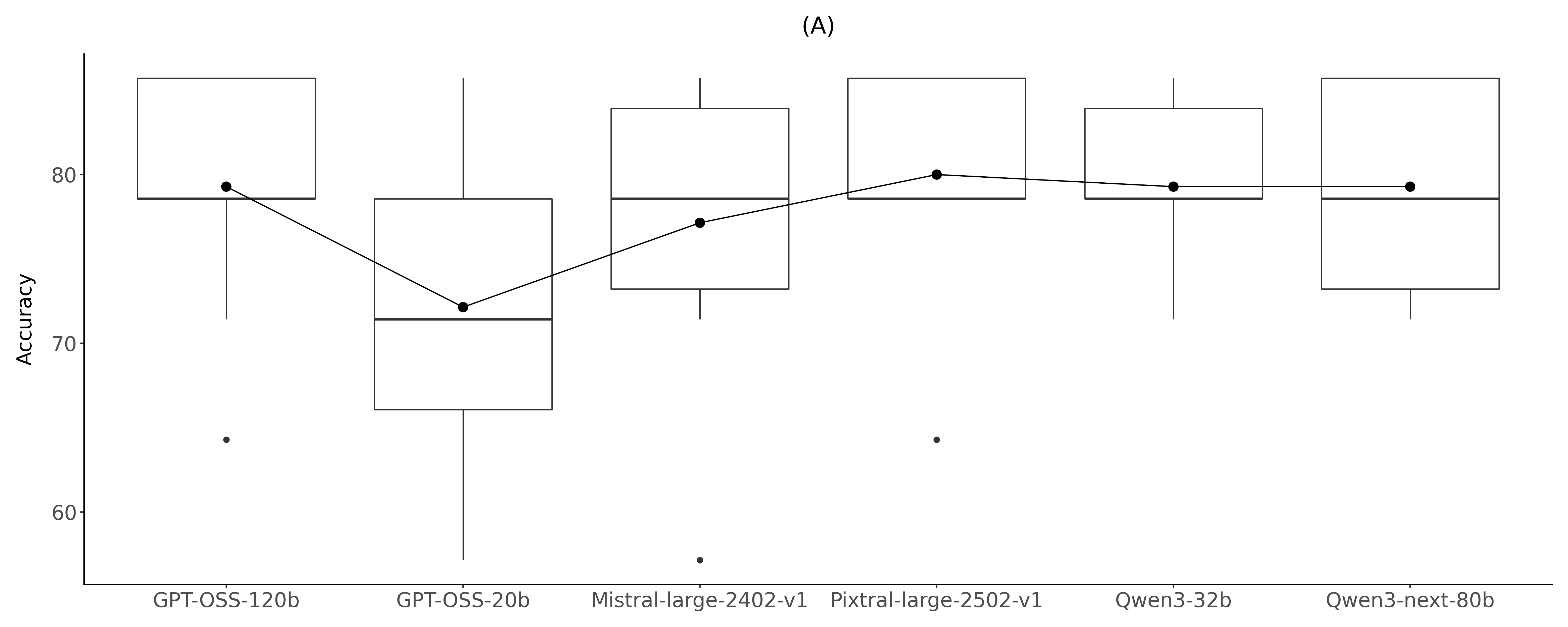}
	\end{subfigure}
	
	\vspace{0.5cm}
	
	\begin{subfigure}[b]{0.85\textwidth}
		\centering
		\includegraphics[width=0.85\linewidth]{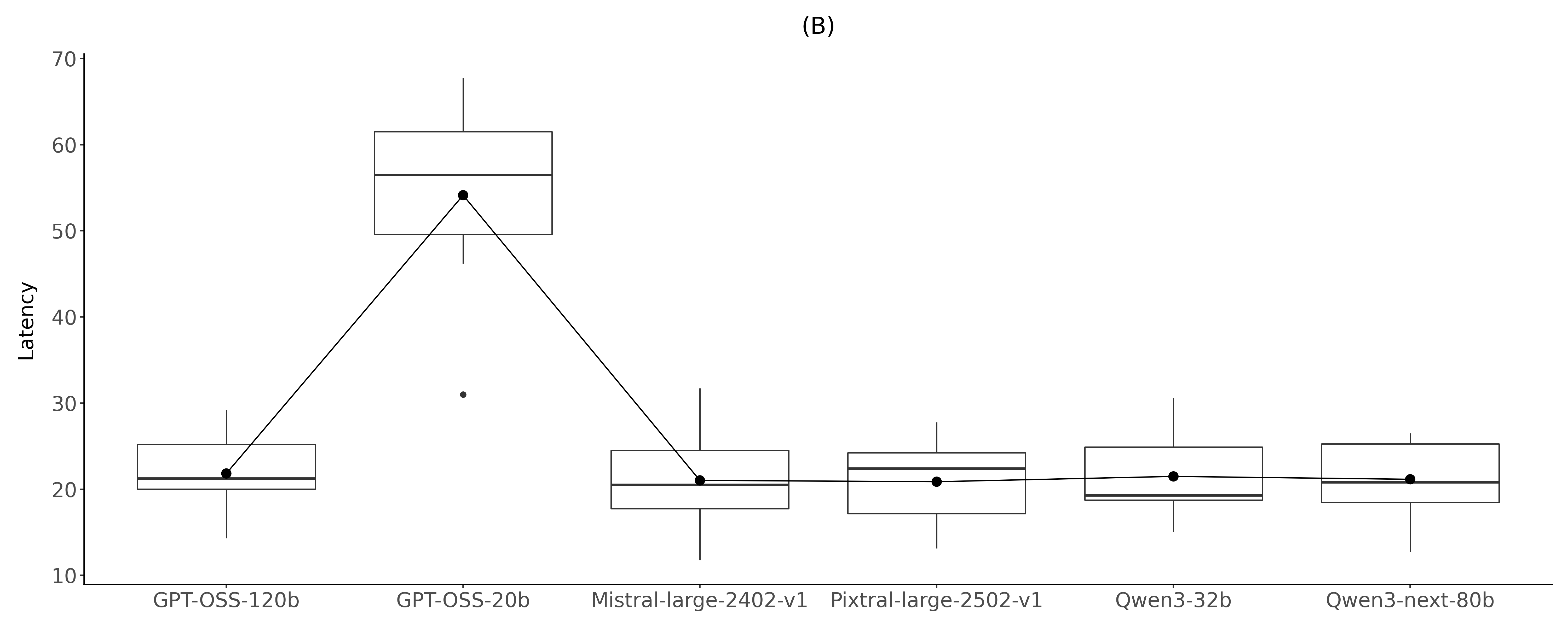}
	\end{subfigure}
	\caption{Boxplots of accuracies in $\%$ (A) and latencies in seconds (B) computed over $N=10$ clinical cases per model. Boxes represent the interquartile range (IQR), the central line indicates the median, and whiskers extend to the most extreme values within $1.5 \times IQR$. Dots denote outliers. The dotted line connects the mean values obtained for each model.}
	\label{fig:fig2}
\end{figure}

Figure \ref{fig:fig2} (A) presents the boxplots of accuracy and the corresponding mean values across the ten lung cancer cases discussed by the tumour board for each evaluated model. As can be observed, the automatic lung cancer form completion system achieves consistently high accuracy levels, thereby demonstrating the practical feasibility of LLM-assisted form autocompletion in the context of multidisciplinary lung cancer tumour committees. Overall, larger models tend to exhibit superior performance. The highest mean accuracies were obtained with the Pixtral-large-2502-v1 model ($80\%$) and with the GPT-OSS-120b, Qwen3-32b, and Qwen3-120b models, all of which reached a mean accuracy of $79.3\%$. In contrast, the lowest mean accuracy was observed for the GPT-OSS-20b model ($72.1\%$). Regarding result stability, the lowest standard deviations were achieved by Qwen3-32b (5.3), Qwen3-80b ($6.3$), Pixtral-large-2502-v1 ($6.6$), and GPT-OSS-120b ($7.1$), indicating more consistent performance across cases, whereas the highest variability was observed for GPT-OSS-20b, with a standard deviation of $10.4$. 

Latency distributions for the evaluated LLMs are shown in Figure \ref{fig:fig2} (B). Two clearly distinct behaviours can be identified. Most models (GPT-OSS-120b, Mistral-large-2402-v1, Pixtral-large-2502-v1, Qwen3-32b and Qwen3-120b) operate within a narrow and homogeneous latency range with mean values concentrated around $20-21$ seconds. In contrast, GPT-OSS-20b exhibits a markedly higher latency, with a mean of $54$ seconds and substantially greater variability. GPT-OSS-120b achieves a latency measure comparable to smaller models, indicating that inference time is driven more by development and serving optimisations than by model size. Mistral family models show the lowest and most stable latency, while Qwen3 variants display similar performance. By contrast, GPT-OSS-20b constitutes an operational outlier and is poorly suited for time-sensitive workflows due to its excessive and unstable response times. 

\section{Conclusions}\label{sec:conclusions}
This work demonstrates the feasibility, robustness and practical relevance of using LLMs to automate the completion of lung cancer tumour board forms within a clinical infrastructure. By integrating a modular data ingestion pipeline, a hybrid relational-vector storage layers and a rule-driven adaptive form model under a RAG architecture, ONCOTIMIA provides a comprehensive and extensible tool for structured clinical documentation powered by GenAI. 

The experimental evaluation across six LLMs provided by AWS Bedrock represents a promising step toward scalable AI-assisted clinical documentation in precision oncology and shows that LLM-assisted autocompletion can achieve high and stable accuracy, approaching $80\%$ correct field completion in tumour board cases. The results also reveal that inference latency is not strictly correlated with model size, since the largest and modern LLMs can deliver response times comparable to smaller systems, making them suitable for clinical use in asynchronous or semi-interactive workflows. From a care perspective, the proposed approach directly addresses one of the main operational bottlenecks in multidisciplinary oncology, and particularly in the implementation of MDTB in hospitals with limited staff resources for the preparation of cases. ONCOTIMIA reduces the manual, repetitive, and error-prone transcription of heterogeneous clinical information into structured forms frequently assigned to clinical staff. By automating a substantial portion of this process, the system has the potential to reduce clinician workload, improve data consistency and accelerate case preparation without altering existing clinical decision workflows. 

Nevertheless, several limitations remain. The current evaluation is based on a limited number of cases and focuses primarily on technical performance metrics. Future work will include larger prospective studies, fine-grained error analysis (e.g., by clinical category) and formal assessment of time savings and user acceptance in real tumour board settings. Finally, further research is needed to strengthen safety guarantees, traceability and explainability, especially in the presence of model hallucinations or incomplete source documentation. 

\section*{Funding sources}
This work is part of project ONCOTIMIA (BAHIA SOFTWARE), that was supported by the IG408M-IA360 program under grant number IG408M-2025-000-000021, funded by the Instituto Galego de Promoción Económica (IGAPE) and cofinanced by the Autonomous Community of Galicia ($25\%$) and the European Union through the Recovery and Resilience Facility ($75\%$), within the framework of the Recovery, Transformation and Resilience Plan-NextGenerationEU, Component 16: National Artificial Intelligence Strategy. 

\bibliographystyle{apalike}
\bibliography{references}

@article{aldea2025,
    author = {Aldea, Mihaela and Rotow, Julia K. and Arcila, Maria and Hatton, Matthew and Sholl, Lynette and Rolfo, Christian and Tagliamento, Marco and Radonic, Teodora and Schalper, Kurt A. and Subbiah, Vivek and Malapelle, Umberto and Roden, Anja C. and Manochakian, Rami and Tsao, Ming-Sound and Linardou, Helena and Hui, Rina and Novello, Silvia and Greystoke, Alastair and Saqi, Anjali and Lantuejoul, Sylvie and Hwang, David M. and Nevins, Kelly and Wynes, Murry and Waqar, Saiama and Han, Yuchen and Yatabe, Yasushi and Chang, Wei-Chin and Hayashi, Takuo and Kim, Tae-Jung and Hofman, Paul and Tavora, Fabio and Hirsch, Fred R. and Denninghoff, Valeria and Leighl, Natasha B. and Drilon, Alexander and Cooper, Wendy A. and Dacic, Sanja and Mohindra, Pranshu and Pavlakis, Nick and Lopez-Rios, Fernando},
    title = {Molecular Tumor Boards: A Consensus Statement From the International Association for the Study of Lung Cancer},
    journal = {Journal of Thoracic Oncology},
    volume = {20},
    number = {11},
    pages = {1594-1614},    
    year = {2025},
    doi = {10.1016/j.jtho.2025.07.009},
    url = {https://doi.org/10.1016/j.jtho.2025.07.009},
}

@article{arndt2017,
    author = {Arndt, B. G. and Beasley, J. W. and Watkinson, M. D. and Temte, J. L. and Tuan, W. J. and Sinsky, C. A. and Gilchrist, V. J.},
    title = {Tethered to the EHR: Primary Care Physician Workload Assessment Using EHR Event Log Data and Time-Motion Observations},
    year = {2017},
    journal = {Annals of family medicine},
    volume = {15(5)},
    pages = {419–426},
    doi = {10.1370/afm.2121},
    url = {https://doi.org/10.1370/afm.2121}
}

@article{ayers2023,
    author = {Ayers, John W. and Poliak, Adam and Dredze, Mark and Leas, Eric C. and Zhu, Zechariah and Kelley, Jessica B. and Faix, Dennis J. and Goodman, Aaron M. and Longhurst, Christopher A. and Hogarth, Michael and Smith, Davey M.},
    title = {Comparing Physician and Artificial Intelligence Chatbot Responses to Patient Questions Posted to a Public Social Media Forum},
    journal = {JAMA Internal Medicine},
    volume = {183},
    number = {6},
    pages = {589-596},
    year = {2023},
    month = {06},
    issn = {2168-6106},
    doi = {10.1001/jamainternmed.2023.1838},
    url = {https://doi.org/10.1001/jamainternmed.2023.1838},
    eprint = {https://jamanetwork.com/journals/jamainternalmedicine/articlepdf/2804309/jamainternal_ayers_2023_oi_230030_1685974538.66672.pdf},
}

@article{bracken2025,
    author = {Aisling Bracken and Clodagh Reilly and Aoife Feeley and Eoin Sheehan and Khalid Merghani and Iain Feeley},
    title = {Artificial Intelligence (AI) – Powered Documentation Systems in Healthcare: A Systematic Review},
    year = {2025},
    journal = {J Med Syst},
    volume = {49},
    pages = {28},
    doi = {10.1007/s10916-025-02157-4},
    url = {https://doi.org/10.1007/s10916-025-02157-4}
}

@article{brown2020,
    author = {Tom B. Brown and
                Benjamin Mann and
                Nick Ryder and
                Melanie Subbiah and
                Jared Kaplan and
                Prafulla Dhariwal and
                Arvind Neelakantan and
                Pranav Shyam and
                Girish Sastry and
                Amanda Askell and
                Sandhini Agarwal and
                Ariel Herbert{-}Voss and
                Gretchen Krueger and
                Tom Henighan and
                Rewon Child and
                Aditya Ramesh and
                Daniel M. Ziegler and
                Jeffrey Wu and
                Clemens Winter and
                Christopher Hesse and
                Mark Chen and
                Eric Sigler and
                Mateusz Litwin and
                Scott Gray and
                Benjamin Chess and
                Jack Clark and
                Christopher Berner and
                Sam McCandlish and
                Alec Radford and
                Ilya Sutskever and
                Dario Amodei},
    title = {Language Models are Few-Shot Learners},
    journal = {CoRR},
    volume = {abs/2005.14165},
    year = {2020},
    url = {https://arxiv.org/abs/2005.14165},
    eprinttype = {arXiv},
    eprint = {2005.14165},
    bibsource = {dblp computer science bibliography, https://dblp.org}
}

@article{chen2025,
    author = {David Chen and Parsa Rod and Swanson Karl and Nunez John-Jose and Critch Andrew and Bitterman Danielle S and Fei-Fei Liu and Srinivas Raman},
    title = {Large language models in oncology: a review},
    journal = {BMJ Oncology},
    volume = {4:e000759},
    year = {2025},
    doi = {10.1136/bmjonc-2025-000759},
    url = {https://doi.org/10.1136/bmjonc-2025-000759},
}

@article{elsaghir2014,
    author = {El Saghir, N. S. and Keating, N. L. and Carlson, R. W. and Khoury, K. E. and Fallowfield, L.},
    title = {Tumor boards: optimizing the structure and improving efficiency of multidisciplinary management of patients with cancer worldwide},
    journal = {American Society of Clinical Oncology educational book. American Society of Clinical Oncology},
    volume = {Annual Meeting},
    pages = {e461–e466},
    year = {2014},
    doi = {10.14694/EdBook_AM.2014.34.e461},
    url = {https://doi.org/10.14694/EdBook_AM.2014.34.e461}
}

@article{esteva2019,
    author = {Andre Esteva and Alexandre Robicquet and Bharath Ramsundar and Volodymyr Kuleshov and Mark DePristo and Katherine Chou and Claire Cui and Greg Corrado and Sebastian Thrun and Jeff Dean},
    title = {A guide to deep learning in healthcare},
    journal = {Nat Med},
    volume = {25},
    pages = {24–29},
    year = {2019},
    doi = {10.1038/s41591-018-0316-z},
    url = {https://doi.org/10.1038/s41591-018-0316-z},
}

@article{gu2021,
    author = {Gu, Yu and Tinn, Robert and Cheng, Hao and Lucas, Michael and Usuyama, Naoto and Liu, Xiaodong and Naumann, Tristan and Gao, Jianfeng and Poon, Hoifung},
    title = {Domain-Specific Language Model Pretraining for Biomedical Natural Language Processing},
    year = {2021},
    issue_date = {January 2022},
    publisher = {Association for Computing Machinery},
    address = {New York, NY, USA},
    volume = {3},
    number = {1},
    url = {https://doi.org/10.1145/3458754},
    doi = {10.1145/3458754},
    journal = {ACM Trans. Comput. Healthcare},
    month = oct,
    articleno = {2},
    numpages = {23}
}

@article{hu2021,
    author = {Hu, D. and Zhang, H. and Li, S. and Wang, Y. and Wu, N. and Lu, X.},
    title = {Automatic Extraction of Lung Cancer Staging Information From Computed Tomography Reports: Deep Learning Approach},
    journal = {JMIR medical informatics},
    volume = {9(7)},
    pages = {e27955},    
    year = {2021},
    doi = {10.2196/27955},
    url = {https://doi.org/10.2196/27955},
}

@article{luo2022,
    author = {Luo, Renqian and Sun, Liai and Xia, Yingce and Qin, Tao and Zhang, Sheng and Poon, Hoifung and Liu, Tie-Yan},
    title = {BioGPT: generative pre-trained transformer for biomedical text generation and mining},
    journal = {Briefings in Bioinformatics},
    volume = {23},
    number = {6},
    pages = {bbac409},
    year = {2022},
    month = {09},
    issn = {1477-4054},
    doi = {10.1093/bib/bbac409},
    url = {https://doi.org/10.1093/bib/bbac409},
    eprint = {https://academic.oup.com/bib/article-pdf/23/6/bbac409/47144271/bbac409.pdf},
}

@article{nori2023,
  title={Capabilities of GPT-4 on Medical Challenge Problems},
  author={Harsha Nori and Nicholas King and Scott Mayer McKinney and Dean Carignan and Eric Horvitz},
  journal={ArXiv},
  year={2023},
  volume={abs/2303.13375},
  url={https://api.semanticscholar.org/CorpusID:257687695}
}

@article{singhal2023,
    author = {Karan Singhal and Shekoofeh Azizi and Tao Tu and S. Sara Mahdavi and Jason Wei and Hyung Won Chung and Nathan Scales and Ajay Tanwani and Heather Cole-Lewis and Stephen Pfohl and Perry Payne and Martin Seneviratne and Paul Gamble and Chris Kelly and Nathaneal Scharli and Aakanksha Chowdhery and Philip Mansfield and Blaise Aguera y Arcas and Dale Webster and Greg S. Corrado and Yossi Matias and Katherine Chou and Juraj Gottweis and Nenad Tomasev and Yun Liu and Alvin Rajkomar and Joelle Barral and Christopher Semturs and Alan Karthikesalingam and Vivek Natarajan},
    title = {Large language models encode clinical knowledge},
    year = {2023},
    issue_date = {August 2023},
    journal = {Nature},
    volume = {620},
    pages = {172–180},
    doi = {10.1038/s41586-023-06291-2},
    url = {https://doi.org/10.1038/s41586-023-06291-2}
}

@article{wang2018,
    author = {Wang, Y. and Wang, L. and Rastegar-Mojarad, M. and Moon, S. and Shen, F. and Afzal, N. and Liu, S. and Zeng, Y. and Mehrabi, S. and Sohn, S. and Liu, H.},
    title = {Clinical information extraction applications: A literature review},
    year = {2018},
    journal = {Journal of biomedical informatics},
    volume = {77},
    pages = {34–49},
    doi = {10.1016/j.jbi.2017.11.011},
    url = {https://doi.org/10.1016/j.jbi.2017.11.011}
}

\end{document}